\documentclass{article}
\usepackage{spconf,amsmath,graphicx}
\usepackage{url}            
\usepackage{booktabs}       
\usepackage{amsfonts}       
\usepackage{nicefrac}       
\usepackage{microtype}      
\usepackage{xcolor}         
\usepackage{times}
\usepackage{soul}
\usepackage{tabularx}
\usepackage{bbm}
\usepackage{dsfont}
\usepackage{wrapfig}
\usepackage{graphicx}
\usepackage{colortbl}
\usepackage{multicol}
\usepackage{makecell}
\usepackage{amsmath,amssymb,amsthm,amsfonts,mathrsfs,bm,multirow}
\usepackage{pifont}
\usepackage{float}
\usepackage[normalem]{ulem}
\usepackage{graphicx}
\usepackage{caption}
\usepackage{subcaption}
\usepackage{makecell}
\usepackage{hyperref}

\usepackage{enumitem}
\usepackage{comment}

\usepackage{algorithmic}
\usepackage{algorithm}
\usepackage{diagbox}

\newcommand{\squishlist}{
    \begin{list}{$\bullet$}
        { \setlength{\itemsep}{0pt}      \setlength{\parsep}{0pt}
            \setlength{\topsep}{0.5pt}       \setlength{\partopsep}{0pt}
            \setlength{\listparindent}{-2pt}
            \setlength{\itemindent}{-5pt}
            \setlength{\leftmargin}{0.5em} \setlength{\labelwidth}{0em}
            \setlength{\labelsep}{0.2em} } }
    
\newcommand{\squishend}{
\end{list}  }

\usepackage[capitalize,noabbrev]{cleveref}
\usepackage[textsize=tiny]{todonotes}
\usepackage{wrapfig}
\usepackage{graphicx}
\usepackage{pifont}

\theoremstyle{plain}

\theoremstyle{definition}

\theoremstyle{remark}

\usepackage{enumitem}
\usepackage{colortbl}

\definecolor{mygray}{gray}{.88}

\title{AutoST: Training-free Neural Architecture Search for Spiking Transformers}
\name{Ziqing Wang, Qidong Zhao, Jinku Cui, Xu Liu, Dongkuan Xu}
\address{North Carolina State University}
\begin{document}
\ninept
\newcommand{\sysname}{\texttt{AutoST} }
\newcommand{\sysnameOnlyAuto}{{\footnotesize{ \texttt{AutoST}}}}
\newcommand{\cmark}{\ding{51}}%
\newcommand{\xmark}{\ding{55}}%
\maketitle
\begin{abstract}
Spiking Transformers have gained considerable attention because they achieve both the energy efficiency of Spiking Neural Networks (SNNs) and the high capacity of Transformers. However, the existing Spiking Transformer architectures, derived from Artificial Neural Networks (ANNs), exhibit a notable architectural gap, resulting in suboptimal performance compared to their ANN counterparts. Manually discovering optimal architectures is time-consuming. To address these limitations, we introduce \sysname, a training-free NAS method for Spiking Transformers, to rapidly identify high-performance Spiking Transformer architectures. Unlike existing training-free NAS methods, which struggle with the non-differentiability and high sparsity inherent in SNNs, we propose to utilize Floating-Point Operations (FLOPs) as a performance metric, which is independent of model computations and training dynamics, leading to a stronger correlation with performance. Our extensive experiments show that \sysname models outperform state-of-the-art manually or automatically designed SNN architectures on static and neuromorphic datasets. Full code, model, and data are released for reproduction.\footnote{\url{https://github.com/AlexandreWANG915/AutoST}}
\end{abstract}
\section{Introduction}
\label{sec:intro}

Spiking neural networks (SNNs) have gained extensive attention owing to their remarkable energy efficiency~\cite{maassNetworksSpikingNeurons1997}. Concurrently, the Transformer has exhibited impressive performance in a wide array of computer vision tasks~\cite{dosovitskiyImageWorth16x162020, liuSwinTransformerHierarchical2021}. This has led to a growing interest in integrating SNNs with Transformers to develop Spiking Transformers, which makes it possible to achieve high energy efficiency and superior performance~\cite{zhouSpikformerWhenSpiking2022, wang_efficient_2022}.

\begin{figure}[t]
\begin{center}
    \includegraphics[width=0.4\textwidth]{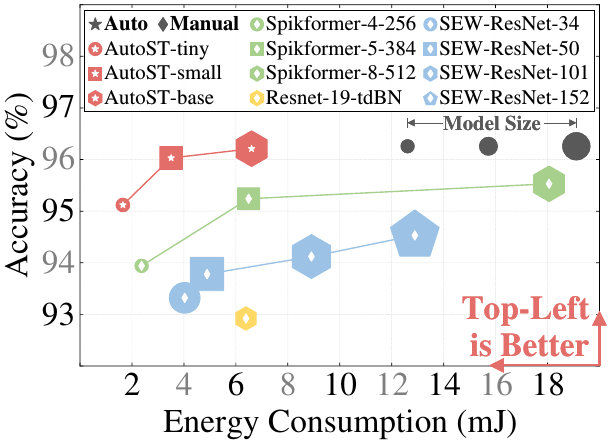}
\caption{Comparison between \sysname and state-of-the-art SNN models on the CIFAR-10 dataset. The marker sizes represent the model sizes. Star-shaped and diamond-shaped hollows within the markers denote manually and automatically designed models, respectively.}
\vskip -0.3in
\label{fig: performance}
\end{center}
\end{figure}

Despite these successes, existing Spiking Transformer architectures predominantly rely on ANN-based architectures. This dependence tends to overlook the unique properties of SNNs, resulting in less optimal performance compared to their ANN counterparts~\cite{kimNeuralArchitectureSearch2022, naAutoSNNEnergyefficientSpiking2022}. As illustrated in Fig.~\ref{fig: performance}, there is a substantial variation in accuracy and energy consumption across different SNN architectures. This variation highlights the need for a more in-depth investigation of the design choices of Spiking Transformer architectures. 

Finding optimal SNN architectures has traditionally been pursued via two common approaches. The first involves manually designing architectures, which is a labor-intensive endeavor~\cite{kimNeuralArchitectureSearch2022} and does not guarantee optimal results~\cite{zhouTrainingfreeTransformerArchitecture2022}. The second approach involves utilizing Neural Architecture Search (NAS) methods to automatically discover the ideal architecture. However, most existing NAS methods require multiple training stages or a single supernet training that encompasses all architecture candidates~\cite{caiProxylessnasDirectNeural2018}, leading to longer convergence times compared to standard training. Considering that the SNNs generally exhibit slower training speeds compared to their ANN counterparts~\cite{kimNeuralArchitectureSearch2022}, the application of these NAS methods to SNNs is challenging.

To address these limitations, we propose, for the first time, to leverage training-free NAS to rapidly identify high-performance Spiking Transformer architectures. Recent training-free NAS approaches~\cite{zhouTrainingfreeTransformerArchitecture2022}, which search for the optimal architecture from initialized networks without any training, substantially reduce search times. Nevertheless, it is not straightforward to directly apply these existing training-free NAS methods to Spiking Transformers. Most current methods, developed for ANNs, rely on gradients during a backward pass; however, the spike is non-differentiable during backpropagation in SNNs, leading to inaccurate gradients. Moreover, methods employing activation patterns introduce significant errors due to the inherent high sparsity of SNNs~\cite{kimNeuralArchitectureSearch2022}.

\begin{figure*}[!t]
\begin{center}
\centerline{\includegraphics[width=0.8\textwidth]{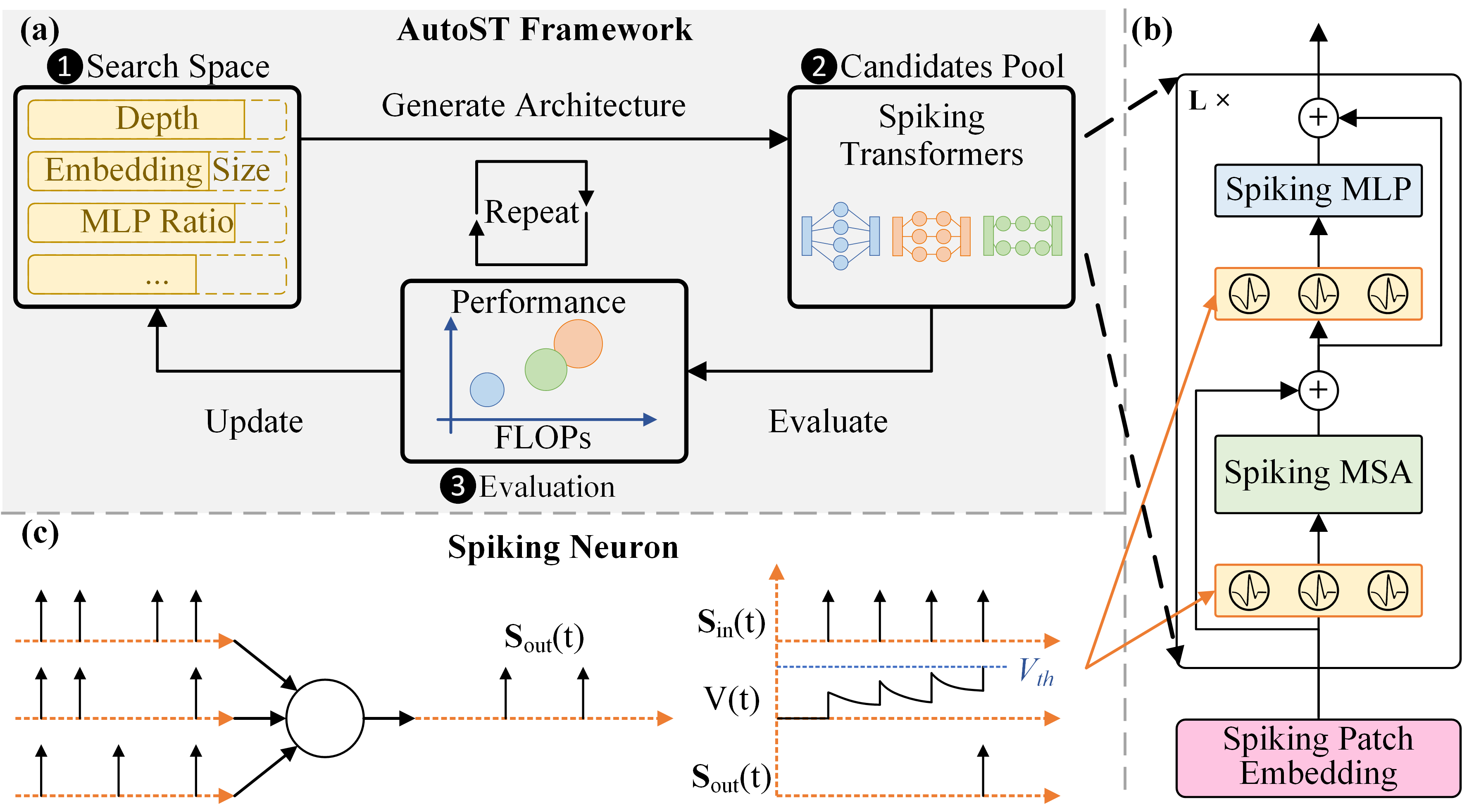}}
\vspace{-0em}
\caption{Overview of \sysname. Subfigure (a) shows the \sysname pipeline: \ding{182} Initial generation of architecture candidates within a predefined search space; \ding{183} Selection of Spiking Transformer architectures from the candidate pool; \ding{184} Evaluation of selected architectures based on proposed training-free metrics. Subfigure (b) presents the overall architecture of Spiking Transformers. Subfigure (c) illustrates the operation of a spiking neuron, which transmits spikes to the subsequent layer when the membrane potential $V(t)$ surpasses the threshold $V_{th}$.}
\label{fig: architecture}
\end{center}
\vspace{-0.35in}
\end{figure*}

In this work, we propose \sysname, a training-free NAS to search for superior Spiking Transformer architectures. Our approach utilizes Floating-Point Operations (FLOPs) as a performance metric, which is independent of model computations and training dynamics, thus effectively tackling the challenges posed by non-differentiability and high sparsity inherent to SNNs, leading to a stronger correlation with performance. To the best of our knowledge, \sysname is the first implementation of training-free NAS to explore Spiking Transformer architectures. Extensive experiments show that our searched \sysname models outperform state-of-the-art SNNs on both static and neuromorphic datasets. In particular, the accuracy of our \sysname model outperforms that of the SNN models searched by other NAS methods by 3.06\%, 4.34\%, and 9.10\% on the CIFAR-10, CIFAR-100, and CIFAR10-DVS, respectively.

\section{Preliminary}
\label{sec:pre}

\subsection{Spiking Neuron Model}
Unlike traditional ANNs, SNNs utilize binary spike trains to convey information. As shown in Fig.~\ref{fig: architecture}, we employ the widely adopted Leaky-Integrate-and-Fire (LIF) model to mimic the dynamics of spiking neurons. The LIF model is defined as follows:
\begin{align}
H[t] &= V[t-1]+\frac{1}{\tau}\left(X[t]-\left(V[t-1]-V_{\text {reset }}\right)\right), \\
S[t] &= \Theta\left(H[t]-V_{t h}\right), \\
V[t] &= H[t](1-S[t])+V_{\text {reset }} S[t],
\end{align}
where $\tau$ denotes the membrane time constant, $X[t]$ represents the input current at the timestep $t$. When the membrane potential $V[t]$ exceeds the firing threshold $V_{t h}$ at the timestep $t$, a spike $S[t]$ will be generated by the spiking neuron to the next layer. The Heaviside step function $\Theta(v)$ equals 1 for $v \geq 0$ and 0 otherwise. $V[t]$ represents the membrane potential post the triggering event, which equals $H[t]$ if no spike is produced and is otherwise reset to the potential $V_{\text {reset }}$.


 \vspace{-0.15em}
\section{Training-free NAS for Spiking Transformers}
\vspace{-0.15em}

\subsection{Performance Prediction via FLOPs}

In this section, we first explore the application of recent training-free metrics for NAS in the context of Spiking Transformers. Many existing metrics require forward and backward passes through the architecture to compute a score, such as SynFlow~\cite{tanakaPruningNeuralNetworks2020a}, Snip~\cite{leeSnipSingleshotNetwork2018} and NTK~\cite{jacotNeuralTangentKernel2018}.

However, SNNs undergo a Heaviside step function during forward propagation, leading to non-differentiability during backward propagation. This intrinsic characteristic of SNNs can result in inaccurate gradient calculations and potentially unreliable metric scores. Furthermore, while the LinearRegions method~\cite{mellorNeuralArchitectureSearch2021} circumvents the need for a backward pass, it faces challenges due to large variations in the sparsity of activation patterns in SNNs~\cite{kimNeuralArchitectureSearch2022}. These variations in sparsity impact the suitability of the LinearRegions method for Spiking Transformers. To overcome these constraints, we propose the use of FLOPs as a metric to predict the final performance of the model. This approach is solely related to the self-characterization of the model and eliminates the need for forward/backward propagation, effectively circumventing both the non-differentiability and sparsity variation issues.

The computation of FLOPs in a Transformer model is predominantly attributed to two components: the Self Attention (SA) block and the Multi-layer Perceptron (MLP) block. We present the computation for each component as follows:
\begin{align}
&\operatorname{FLOPs}_{\text{MSA}} = L \times n \times d_{model} \times (2d_{model} + n)
\label{eq:self-attention} \\
&\operatorname{FLOPs}_{\text{MLP}} = L \times n \times (d_{model} \times d_{mlp} + d_{mlp} \times d_{model})
\label{eq:ffn}
\end{align}

For the SA block, $\operatorname{FLOPs}_{\text{SA}}$ denotes the total number of floating-point operations required, where $L$ represents the number of layers, $n$ the sequence length, and $d_{model}$ is the embedding dimension of the model. For the MLP block, $d_{mlp}$ represents the hidden dimension.

\begin{figure}
\begin{center}
    \includegraphics[width=0.25\textwidth]{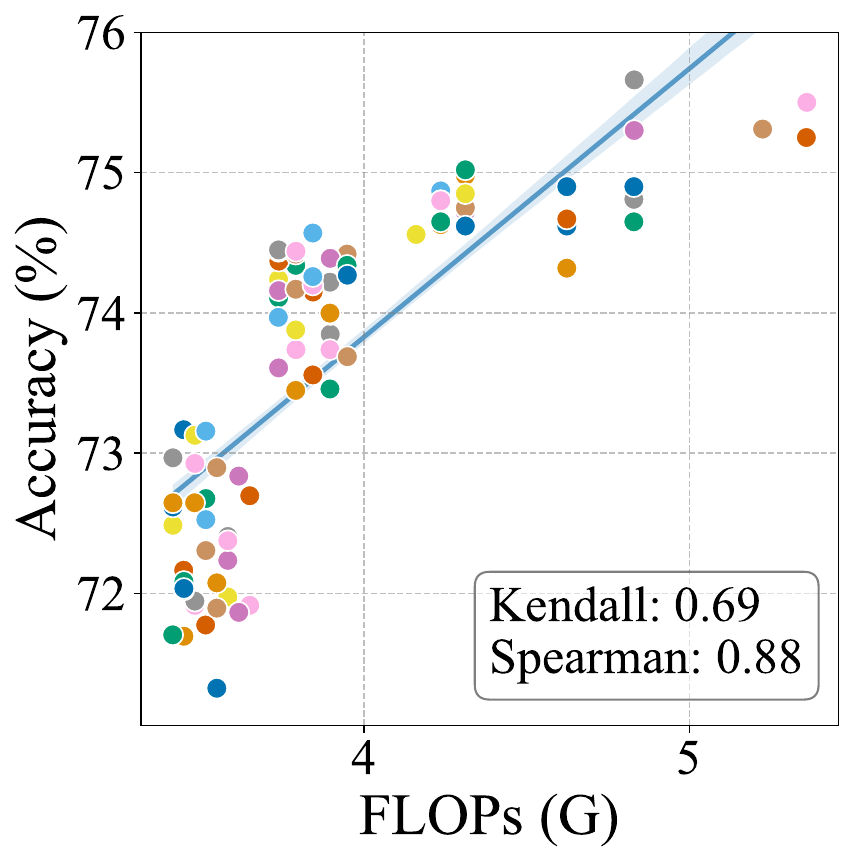}
\vskip -0.05in
\caption{Performance Metric Evaluation: Evidence of a significant positive correlation between FLOPs and accuracy.}
\vskip -0.15in
\label{flops}
\end{center}
\end{figure}

\begin{table}[!t]
\setlength\tabcolsep{2.5pt} 
\renewcommand{\arraystretch}{0.9}
\begin{center}
\begin{small}
\caption{Comparison of Correlation Coefficients: Analysis of the proposed metric against different training-free metrics. Note that absolute values of Kendall and Spearman coefficients approaching one indicate a stronger correlation between two variables. The search time represents the duration required to search 200 samples on the CIFAR-100 dataset with a single GPU. Our metric exhibits a pronounced correlation with the final accuracy.}
\label{spearman}
\vskip -0.1in
\begin{tabular}{c*{2}{c}*{2}{c}{c}}
\toprule \toprule
\textbf{Datasets}      & \multicolumn{2}{c}{\textbf{CIFAR-10}} & \multicolumn{2}{c}{\textbf{CIFAR-100}} &  \\ \midrule
\textbf{Metric}        & \textbf{Kendall}  & \textbf{Spearman} & \textbf{Kendall}  & \textbf{Spearman}  & \\ \midrule
\textbf{NTK~\cite{jacotNeuralTangentKernel2018}}           & -0.43                  & -0.49                  &   -0.40                &  -0.55                          \\
\textbf{Snyflow~\cite{tanakaPruningNeuralNetworks2020a}}       &  0.57                 &   0.77                &    0.56               &  0.76                      \\
\textbf{LinearRegions~\cite{mellorNeuralArchitectureSearch2021}} &  0.58            &  0.85                 & 0.59              & 0.79                  \\
\textbf{SAHD~\cite{kimNeuralArchitectureSearch2022}}          &  0.56             &  0.81                 & 0.52              & 0.74                 \\
\rowcolor{mygray}\textbf{\sysname (ours)}  &  \textbf{0.70}       &  \textbf{0.91}                 & \textbf{0.69}     & \textbf{0.88}      
\\ \hline     
\bottomrule
\end{tabular}
\end{small}
\end{center}
\vskip -0.3in
\end{table}

The computation of the FLOPs of Spiking Transformers is similar to the process for ANN Transformers, with additional consideration for the time dimension. To account for this, we multiply the FLOPs of the ANN-equivalent architecture by the number of timesteps: $\operatorname{FLOPs}_{\text{SNN}} = T \times \operatorname{FLOPs}_{\text{ANN}}$.

To demonstrate the effectiveness of our proposed metric, \sysname, we randomly sample 100 architectures from the small search space outlined in Tab.~\ref{tab: search space}. We initially train these architectures for 100 epochs, compute the training-free metric scores, and then analyze their correlation with the final model performance. As shown in Tab.~\ref{spearman} and Fig.~\ref{flops}, our \sysname has high absolute values for both Kendall and Spearman coefficients. This suggests that the FLOPs metric has a strong correlation with model performance. This correlation is largely due to two factors. First, the FLOPs metric is independent of gradients, which allows it to circumvent the inherent non-differentiability of SNNs. Second, previous works assessed their metrics over a large parameter range, whereas, real-world applications often demand the identification of the optimal model within specific constraints (e.g., 4-5 M). In such circumstances, traditional metrics struggle to distinguish model differences in such a narrow parameter range. In contrast, \sysname using FLOPs can effectively distinguish different models due to its calculation based on layer numbers and embedding dimensions.

\vskip -0.5in
\subsection{Overall Architecture of \sysname} \label{Overall Architecture}
The overall architecture of \sysname is based on~\cite{zhouSpikformerWhenSpiking2022}, a purely transformer-based SNN. The overview of \sysname is depicted in Fig.~\ref{fig: architecture}. Considering an input 2D image sequence $I \in \mathbb{R}^{T \times C \times H \times W}$, a Spiking Patch Embedding (SPE) block is utilized for downsampling and partitioning the input into spiking patches $X \in \mathbb{R}^{T \times N \times D}$. These patches are subsequently passed through $L$ Spiking Transformer Blocks, each comprising a Spiking Self Attention (SSA) block and a Spiking MLP (SMLP) block. The final stage consists of a Global Average Pooling (GAP) and a Fully Connected Layer (FC) serving as the classification head, outputting the prediction $Y$. The sequence of operations is mathematically defined as follows:
\begin{align}
&X=\operatorname{SPE}(I), \quad I \in \mathbb{R}^{T \times C \times H \times W}, \quad  X \in \mathbb{R}^{T \times N \times D} \\
&X_l^{\prime}=\operatorname{SSA}\left(X_{l-1}\right)+X_{l-1}, \quad X_l^{\prime} \in \mathbb{R}^{T \times N \times D}, \quad l=1 \ldots L \\
&X_l=\operatorname{SMLP}\left(X_l^{\prime}\right)+X_l^{\prime}, \quad X_l \in \mathbb{R}^{T \times N \times D}, \quad l=1 \ldots L \\
&Y=\operatorname{FC}\left(\operatorname{GAP}\left(X_L\right)\right)
\end{align}
\subsection{Search Space and Search Algorithm of \sysname} \label{search space}

We design an extensive search space for \sysname, comprising four key variables in the Transformer: embedding size, number of heads, MLP ratio, and network depth, as detailed in Tab.~\ref{tab: search space}. Based on these five variables, we design three distinct search spaces (Tiny, Small, and Base) for \sysname. Given the constraints on the model parameters, we divide the large-scale search space into three parts, as described in Tab.~\ref{tab: search space}. This partitioning enables the search algorithm to focus on discovering models within a specific parameter range. And we conduct an evolutionary search to obtain the optimal \sysname under specified resource constraints.

\begin{table}[h]
  \setlength\tabcolsep{1.5pt}
  \renewcommand{\arraystretch}{0.9}
  \begin{center}
    \begin{small}
      \vskip -0.1in
      \caption{The search space of \sysname. Tuples in parentheses represent the lowest, highest value, and steps.}
        \label{tab: search space}
      \begin{tabular}{cccc}
        \toprule \toprule
                        & \textbf{\sysname-tiny} & \textbf{\sysname-small} & \textbf{\sysname-base} \\ \midrule
        \textbf{Embed Size}    & (192, 384, 64)    & (256, 512, 64)     & (384, 768, 64)    \\
        \textbf{MLP Ratio}    & (3, 5, 1)      & (3, 5, 1)       & (3, 6, 1)      \\
        \textbf{Head Num}     & (4, 8, 4)         & (4, 8, 4)          & (4, 8,4)          \\
        \textbf{Depth}    & (1,8,1)           & (2,12,1)           & (4,15,1)          \\ 
        \textbf{\# Params} & 4-5M              & 11-15M             & 25-35M           
      \\ \hline     
      \bottomrule
      \end{tabular}
    \end{small}
  \end{center}
  \vskip -0.35in
\end{table}


\begin{table*}[!t]
\setlength\tabcolsep{3pt} 
\renewcommand{\arraystretch}{0.9}
\begin{center}
\begin{small}
\caption{Performance comparison between the proposed \sysname model and the state-of-the-art models on the CIFAR-10 and CIFAR-100 datasets. * represent the result of our implementation. Acc. denotes the top-1 accuracy.}
\label{model_comparison}
\vskip -0.1in
\begin{tabular}{c c c c c c c}
\toprule \toprule
\textbf{Methods} & \textbf{\# Param (M)} & \textbf{Timesteps} & \textbf{CIFAR-10 Acc. (\%)} & \textbf{CIFAR-100 Acc. (\%)} & \textbf{Model Type} & \textbf{Design Type} \\
\midrule
 Spikformer-4-256~\cite{zhouSpikformerWhenSpiking2022}~[ICLR23] & 4.15 & 4 & 93.94 & 75.96 & Transformer & Manual \\
  AutoSNN~\cite{miaoNeuromorphicVisionDatasets2019}~[ICML22] & 5.44 & 8 & 92.54 & 69.16 & CNN & Auto \\
 SNASNet-Bw~\cite{buOptimalANNSNNConversion2021a}~[ECCV22] & - & 5 & 93.64 & 73.04 & CNN & Auto \\
 \cellcolor{mygray}\sysname-tiny (ours) & \cellcolor{mygray}4.20 & \cellcolor{mygray}4 & \cellcolor{mygray}\textbf{95.14} & \cellcolor{mygray}\textbf{76.29} & \cellcolor{mygray}Transformer & \cellcolor{mygray}Auto \\
\midrule
  TET~\cite{dengTemporalEfficientTraining2022}~[ICLR22] & 12.60 & 6 & 94.50 & 74.72 & CNN & Manual \\
  DSR~\cite{mengTrainingHighPerformanceLowLatency2022}~[CVPR22] & 11.20 & 20 & 95.40 & 78.20 & CNN & Manual \\
 Spikformer-5-384*~\cite{zhouSpikformerWhenSpiking2022}~[ICLR23] & 11.32 & 4 & 95.24 & 78.12 & Transformer & Manual \\
  SpikeDHS~\cite{miaoNeuromorphicVisionDatasets2019}~[CVPR23] & 12.00 & 6 & 94.34 & 75.70 & CNN & Auto \\
  \cellcolor{mygray}\sysname-small (ours) & \cellcolor{mygray}11.52 & \cellcolor{mygray}4 & \cellcolor{mygray}\textbf{96.03} & \cellcolor{mygray}\textbf{79.44} & \cellcolor{mygray}Transformer & \cellcolor{mygray}Auto \\
\midrule
  Diet-SNN~\cite{rathiDIETSNNDirectInput2020a}~[TNNLS21] & 39.90 & 5 & 93.44 & 69.67 & CNN & Manual \\
  RMP~\cite{hanRmpsnnResidualMembrane2020a}~[CVPR20] & 39.90 & 2048 & 93.63 & 70.93 & CNN & Manual \\
  Spikformer-8-512*~\cite{zhouSpikformerWhenSpiking2022}~[ICLR23] & 29.68 & 4 & 95.53 & 78.48 & Transformer & Manual \\
  \cellcolor{mygray}\sysname-base (ours) & \cellcolor{mygray}29.64 & \cellcolor{mygray}4 & \cellcolor{mygray}\textbf{96.21} & \cellcolor{mygray}\textbf{79.69} & \cellcolor{mygray}Transformer & \cellcolor{mygray}Auto \\ \hline
\bottomrule
\end{tabular}
\vskip -0.8in
\end{small}
\end{center}
\end{table*}

\vskip -0.4in
\section{Experiments}
\subsection{Performance on CIFAR Datasets}

\textbf{Tiny Models.} \sysname surpasses both Spikformer~\cite{zhouSpikformerWhenSpiking2022} and AutoSNN~\cite{miaoNeuromorphicVisionDatasets2019}. Specifically, the proposed \sysname surpasses AutoSNN, which is obtained through NAS, by 2.6\%, 10.1\%, and 13.01\% on three datasets with fewer parameters and timesteps. This indicates that our model leverages the efficient search space of the \sysname, allowing for enhanced performance with reduced model~{complexity.} \textbf{Small Models.} The larger search space provides additional flexibility, potentially enhancing performance. Our \sysname outperforms Spikformer~\cite{zhouSpikformerWhenSpiking2022} in the same parameter range and yields higher accuracy compared to CNN-based models. Remarkably, \sysname not only outperforms the current state-of-the-art SNN model, DSR~\cite{mengTrainingHighPerformanceLowLatency2022}, in terms of accuracy but also demonstrates significantly fewer time steps, leading to lower SNN latency. This performance suggests that our \sysname method of utilizing an expanded search space can lead to significant performance gains in SNN~{architectures.} \textbf{Base Models.} \sysname maintains its superior performance, outperforming all other models with substantially fewer parameters in a larger search space. This further highlights the effectiveness and efficiency of our proposed method, potentially opening new avenues for SNN.

\subsection{Performance on ImageNet Dataset}

In Tab.~\ref{tab: imagenet}, we showcase the performance of several models on the ImageNet dataset. The SEW ResNet~\cite{zhengGoingDeeperDirectlytrained2021} model has two variations: SEW-ResNet-50 with 25.56M parameters (67.04\% accuracy) and SEW-ResNet-101 with 44.55M parameters (67.78\% accuracy). The Spikformer~\cite{zhouSpikformerWhenSpiking2022} model offers configurations like Spikformer-6-256 (5.99M parameters, 62.99\% accuracy) and scaled-up versions Spikformer-8-384 and Spikformer-10-512, achieving accuracies of 70.24\% and 73.68\% respectively. Our proposed AST model, \sysname, demonstrates a competitive performance-to-parameter ratio. Notably, \sysname-small, with only 14.68M parameters, outperforms Spikformer-8-384 with an accuracy of 71.02\%. The \sysname-base configuration achieves the highest accuracy of 74.54\% among all models in the table with its 34.44M parameters. In essence, \sysname delivers outstanding performance on large-scale datasets.

\begin{table}[t]
\setlength\tabcolsep{2pt} 
\renewcommand{\arraystretch}{0.9}
\vskip -0.25in
\begin{small}
\caption{Performance comparison on the ImageNet dataset.}
\label{tab: imagenet}
\begin{tabular}{ccccc}
\toprule \toprule
\textbf{Methods}                     & \textbf{\# Param (M)} & \textbf{Ts}          & \textbf{Architecture}      & \textbf{Acc (\%)}            \\ \midrule
\multirow{2}{*}{\textbf{SEW ResNet}~\cite{zhengGoingDeeperDirectlytrained2021}} & 25.56    & \multirow{2}{*}{6} & SEW-ResNet-50     & 67.04          \\
                            & 44.55    &                    & SEW-ResNet-101    & 67.78          \\ \hline
\multirow{3}{*}{\textbf{Spikfomer}~\cite{zhouSpikformerWhenSpiking2022} }  & 5.99     & \multirow{3}{*}{4} & Spikformer-6-256  & 62.99          \\ 
                            & 16.81    &                    & Spikformer-8-384  & 70.24          \\
                            & 29.68    &                    & Spikformer-10-512 & 73.68          \\ \hline
\multirow{3}{*}{\textbf{\sysname (ours)}}        & 5.99     & \multirow{3}{*}{4} & \sysname-tiny          & 63.80          \\ 
                            & 14.68    &                    & \sysname-small         & 71.02          \\
                            & 34.44    &                    & \sysname-base          & \textbf{74.54}
\\\hline
\bottomrule
\end{tabular}
\end{small}
\vskip -0.1in
\end{table}

\begin{table}[h]
\setlength\tabcolsep{1.5pt} 
\renewcommand{\arraystretch}{0.9}
\begin{small}
\caption{Performance comparison on the CIFAR10-DVS dataset.}
\label{tab: neuromorphic}
\begin{tabular}{ccccc}
\toprule \toprule
\textbf{Methods}              & \textbf{Timesteps} & \textbf{Acc (\%)} & \textbf{Model Type} & \textbf{Design Type} \\ \midrule
\textbf{tdBN}~\cite{zhengGoingDeeperDirectlytrained2021}                 & 10                 & 67.8                   & CNN                 & Manual               \\
\textbf{LIAF-Net}~\cite{wuLiafnetLeakyIntegrate2021}             & 10                 & 70.4                   & CNN                 & Manual               \\
\textbf{PLIF}~\cite{fangIncorporatingLearnableMembrane2021}                 & 20                 & 74.8                   & CNN                 & Manual               \\
\textbf{Dspkie}~\cite{liDifferentiableSpikeRethinking2021}               & 10                 & 75.4                   & CNN                 & Manual               \\
\textbf{DSR}~\cite{mengTrainingHighPerformanceLowLatency2022}                  & 10                 & 77.3                   & CNN                 & Manual               \\
\textbf{SEW-ResNet}~\cite{zhengGoingDeeperDirectlytrained2021}          & 16                 & 74.4                   & CNN                 & Manual               \\
\textbf{Spikformer}~\cite{zhouSpikformerWhenSpiking2022}           & 16                 & 80.9                   & Transformer         & Manual               \\
\textbf{AutoSNN}~\cite{miaoNeuromorphicVisionDatasets2019}              & 8                  & 72.5                   & CNN                 & Auto                 \\
\rowcolor{mygray}\textbf{\sysname (ours)} & 16                 & \textbf{81.6}          & Transformer         & Auto                
\\\hline
\bottomrule
\end{tabular}
\end{small}
\vskip -0.2in
\end{table}

\subsection{Performance on Neuromorphic Datasets}

We further demonstrate the superiority of the \sysname method by evaluating its performance on neuromorphic datasets. We compare our model with state-of-the-art SNN models on the CIFAR10-DVS dataset. Due to the overfitting issue on neuromorphic datasets, we restrict our comparison to the tiny model. Tab.~\ref{tab: neuromorphic} reveals that the \sysname model achieves the highest top-1 accuracy, outperforming the existing state-of-the-art SNN models. Notably, \sysname outperforms SEW-ResNet~\cite{zhengGoingDeeperDirectlytrained2021} by 7.2\% while maintaining far fewer parameters. These results further validate the effectiveness of our proposed method and its strong generalization capabilities when applied to neuromorphic datasets.



\begin{figure}[t]
     \centering
\vskip -0.2in   
     \begin{subfigure}[b]{0.15\textwidth}
         \centering
         \includegraphics[width=\textwidth]{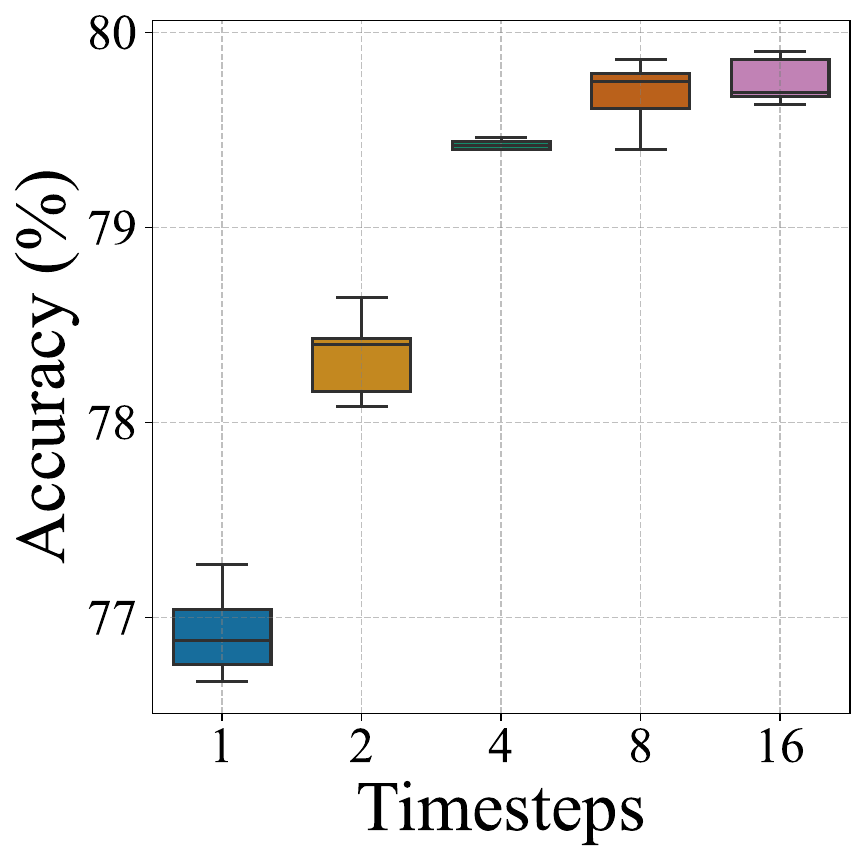}
         \caption{Timesteps \& Acc}
         \label{fig:timesteps-accuracy}
     \end{subfigure}
     \hfill
     \begin{subfigure}[b]{0.15\textwidth}
         \centering
         \includegraphics[width=\textwidth]{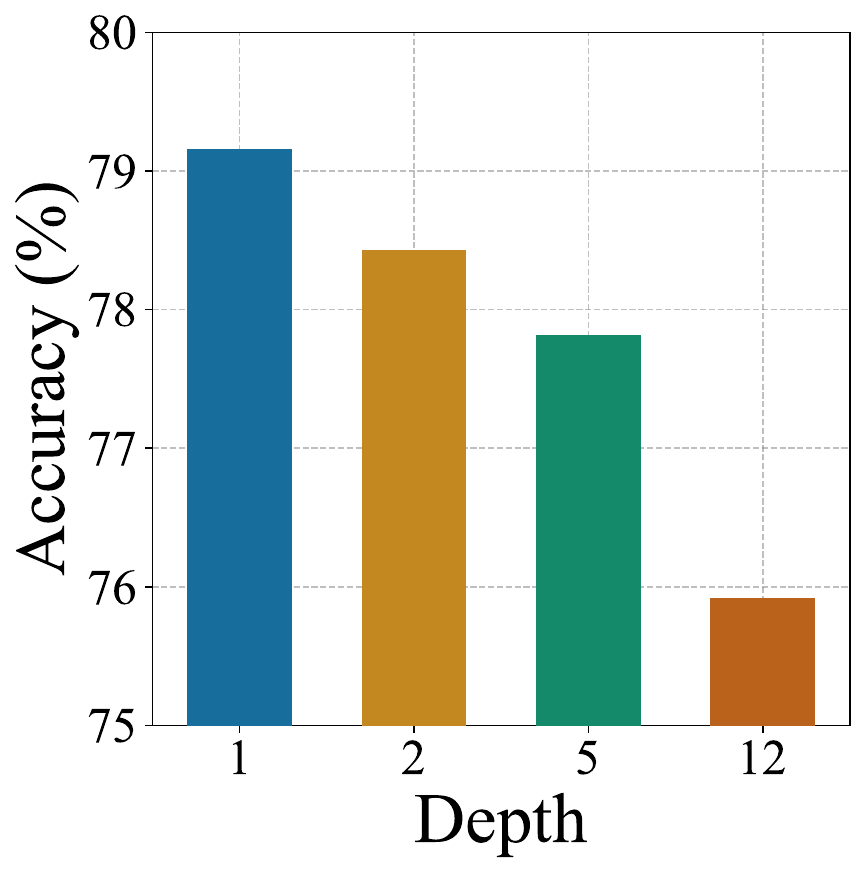}
         \caption{Depth \& Acc}
         \label{fig:depth-accuracy}
     \end{subfigure}
     \hfill
    \begin{subfigure}[b]{0.15\textwidth}
         \centering
         \includegraphics[width=\textwidth]{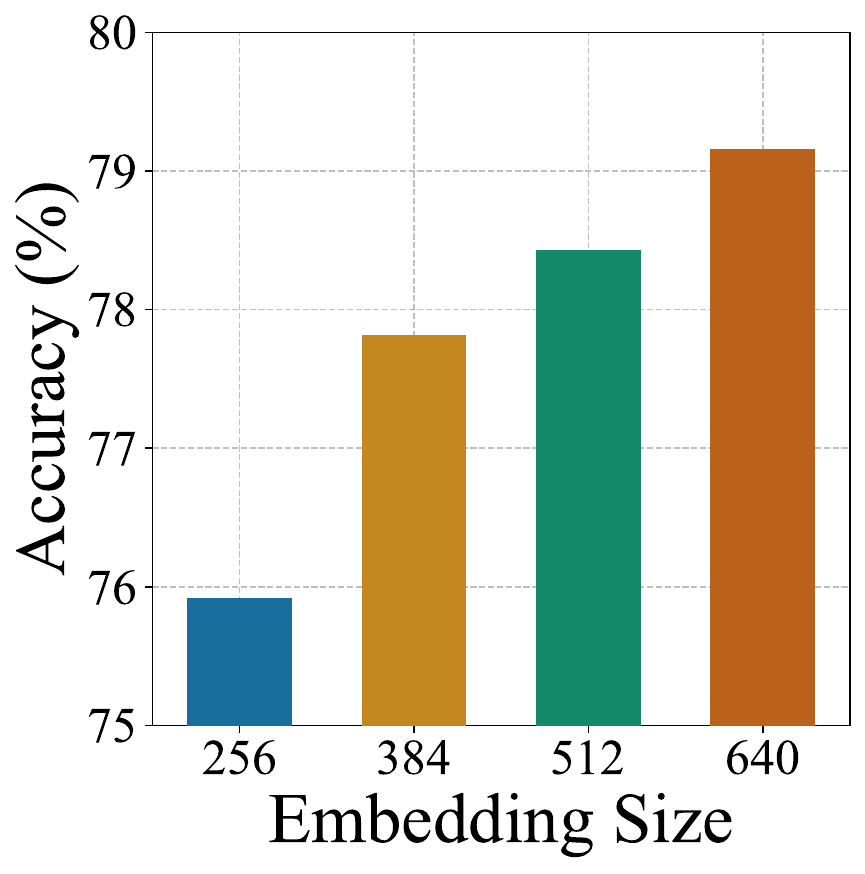}
         \caption{Embed \& Acc}
         \label{fig:embedsize-accuracy}
     \end{subfigure}
        \caption{Subfigure (a) reveals that timesteps and accuracy have a positive correlation. Subfigure (a-c) highlights the advantages of the search \sysname architectures: higher accuracy, lower training time, increased FLOPs, and decreased power consumption. }   
        \label{fig:four-graphs}
\vskip -0.2in   
\end{figure}

\subsection{Analysis of Model Topology and Timestep Optimization}

\textbf{Timestep Optimization.} As depicted in Fig.~\ref{fig:timesteps-accuracy}, there is a discernible positive correlation between the number of timesteps and the performance of the model. While an increase in timesteps tends to enhance accuracy, it concomitantly leads to greater latency and energy consumption. Through our analysis, a timestep value of 4 emerges as a pivotal point that amplifies accuracy without compromising efficiency, thus serving as our preferred choice for subsequent experiments.

\textbf{Model Topology Insights.} Our exploration into the search rule of \sysname has unveiled an inclination towards network structures that are shallower yet broader, marking a distinct shift from the conventional preference of SNNs for configurations that are deeper and narrower~\cite{zhengGoingDeeperDirectlyTrained2020}. To validate this observation, we manually crafted and assessed four distinct architectures, varying in embedding sizes and depths but within a similar parameter range. Figures~\ref{fig:depth-accuracy}-\ref{fig:embedsize-accuracy} corroborate our hypothesis, indicating that architectures with increased width consistently outperform their counterparts in terms of accuracy. This phenomenon is likely attributable to the intrinsic characteristics of SNNs. Specifically, architectures that are broader but not as deep can mitigate the accumulation of quantization errors across layers~\cite{buOptimalANNSNNConversion2021}. This observed predilection for broader Transformers aligns well with recent scholarly insights~\cite{zhaiScalingVisionTransformers2022}.


\vspace{-0.15em}
\section{Conclusion}
\vspace{-0.15em}

In this paper,we introduce \sysname, a training-free NAS method for Spiking Transformers. Our unique approach of utilizing Floating-Point Operations (FLOPs) as a performance metric bypasses the non-differentiability and high sparsity issues present in SNNs. Empirical results confirm that \sysname models consistently outperform both manual and other automatic SNN architectures across diverse datasets. In essence, \sysname paves the way for optimizing Spiking Transformers, promising enhanced performance and efficiency.



\newpage
{\small
\bibliographystyle{IEEEbib}
\bibliography{reference}

\begin{thebibliography}{10}

\bibitem{maassNetworksSpikingNeurons1997}
Wolfgang Maass,
\newblock ``Networks of spiking neurons: The third generation of neural network models,''
\newblock {\em Neural networks}, vol. 10, no. 9, pp. 1659--1671, 1997.

\bibitem{dosovitskiyImageWorth16x162020}
Alexey Dosovitskiy, Lucas Beyer, Alexander Kolesnikov, Dirk Weissenborn, Xiaohua Zhai, Thomas Unterthiner, Mostafa Dehghani, Matthias Minderer, Georg Heigold, and Sylvain Gelly,
\newblock ``An image is worth 16x16 words: {{Transformers}} for image recognition at scale,''
\newblock {\em arXiv preprint arXiv:2010.11929}, 2020.

\bibitem{liuSwinTransformerHierarchical2021}
Ze~Liu, Yutong Lin, Yue Cao, Han Hu, Yixuan Wei, Zheng Zhang, Stephen Lin, and Baining Guo,
\newblock ``Swin transformer: {{Hierarchical}} vision transformer using shifted windows,''
\newblock in {\em Proceedings of the {{IEEE}}/{{CVF International Conference}} on {{Computer Vision}}}, 2021, pp. 10012--10022.

\bibitem{zhouSpikformerWhenSpiking2022}
Zhaokun Zhou, Yuesheng Zhu, Chao He, Yaowei Wang, Shuicheng Yan, Yonghong Tian, and Li~Yuan,
\newblock ``Spikformer: {{When Spiking Neural Network Meets Transformer}},''
\newblock {\em arXiv preprint arXiv:2209.15425}, 2022.

\bibitem{wang_efficient_2022}
Ziqing Wang, Yuetong Fang, Jiahang Cao, Zhongrui Wang, and Renjing Xu,
\newblock ``Efficient {Spiking} {Transformer} {Enabled} {By} {Partial} {Information},'' Oct. 2022,
\newblock arXiv:2210.01208 [cs].

\bibitem{kimNeuralArchitectureSearch2022}
Youngeun Kim, Yuhang Li, Hyoungseob Park, Yeshwanth Venkatesha, and Priyadarshini Panda,
\newblock ``Neural architecture search for spiking neural networks,''
\newblock in {\em Computer {{Vision}}\textendash{{ECCV}} 2022: 17th {{European Conference}}, {{Tel Aviv}}, {{Israel}}, {{October}} 23\textendash 27, 2022, {{Proceedings}}, {{Part XXIV}}}. 2022, pp. 36--56, {Springer}.

\bibitem{naAutoSNNEnergyefficientSpiking2022}
Byunggook Na, Jisoo Mok, Seongsik Park, Dongjin Lee, Hyeokjun Choe, and Sungroh Yoon,
\newblock ``{{AutoSNN}}: Towards energy-efficient spiking neural networks,''
\newblock in {\em International {{Conference}} on {{Machine Learning}}}. 2022, pp. 16253--16269, {PMLR}.

\bibitem{zhouTrainingfreeTransformerArchitecture2022}
Qinqin Zhou, Kekai Sheng, Xiawu Zheng, Ke~Li, Xing Sun, Yonghong Tian, Jie Chen, and Rongrong Ji,
\newblock ``Training-free transformer architecture search,''
\newblock in {\em Proceedings of the {{IEEE}}/{{CVF Conference}} on {{Computer Vision}} and {{Pattern Recognition}}}, 2022, pp. 10894--10903.

\bibitem{caiProxylessnasDirectNeural2018}
Han Cai, Ligeng Zhu, and Song Han,
\newblock ``Proxylessnas: {{Direct}} neural architecture search on target task and hardware,''
\newblock {\em arXiv preprint arXiv:1812.00332}, 2018.

\bibitem{tanakaPruningNeuralNetworks2020a}
Hidenori Tanaka, Daniel Kunin, Daniel~L. Yamins, and Surya Ganguli,
\newblock ``Pruning neural networks without any data by iteratively conserving synaptic flow,''
\newblock {\em Advances in neural information processing systems}, vol. 33, pp. 6377--6389, 2020.

\bibitem{leeSnipSingleshotNetwork2018}
Namhoon Lee, Thalaiyasingam Ajanthan, and Philip~HS Torr,
\newblock ``Snip: {{Single-shot}} network pruning based on connection sensitivity,''
\newblock {\em arXiv preprint arXiv:1810.02340}, 2018.

\bibitem{jacotNeuralTangentKernel2018}
Arthur Jacot, Franck Gabriel, and Cl{\'e}ment Hongler,
\newblock ``Neural tangent kernel: {{Convergence}} and generalization in neural networks,''
\newblock {\em Advances in neural information processing systems}, vol. 31, 2018.

\bibitem{mellorNeuralArchitectureSearch2021}
Joe Mellor, Jack Turner, Amos Storkey, and Elliot~J. Crowley,
\newblock ``Neural architecture search without training,''
\newblock in {\em International {{Conference}} on {{Machine Learning}}}. 2021, pp. 7588--7598, {PMLR}.

\bibitem{miaoNeuromorphicVisionDatasets2019}
Shu Miao, Guang Chen, Xiangyu Ning, Yang Zi, Kejia Ren, Zhenshan Bing, and Alois Knoll,
\newblock ``Neuromorphic vision datasets for pedestrian detection, action recognition, and fall detection,''
\newblock {\em Frontiers in neurorobotics}, vol. 13, pp. 38, 2019.

\bibitem{buOptimalANNSNNConversion2021a}
Tong Bu, Wei Fang, Jianhao Ding, PengLin Dai, Zhaofei Yu, and Tiejun Huang,
\newblock ``Optimal {{ANN-SNN Conversion}} for {{High-accuracy}} and {{Ultra-low-latency Spiking Neural Networks}},''
\newblock in {\em International {{Conference}} on {{Learning Representations}}}, 2021.

\bibitem{dengTemporalEfficientTraining2022}
Shikuang Deng, Yuhang Li, Shanghang Zhang, and Shi Gu,
\newblock ``Temporal {{Efficient Training}} of {{Spiking Neural Network}} via {{Gradient Re-weighting}},''
\newblock {\em arXiv preprint arXiv:2202.11946}, 2022.

\bibitem{mengTrainingHighPerformanceLowLatency2022}
Qingyan Meng, Mingqing Xiao, Shen Yan, Yisen Wang, Zhouchen Lin, and Zhi-Quan Luo,
\newblock ``Training {{High-Performance Low-Latency Spiking Neural Networks}} by {{Differentiation}} on {{Spike Representation}},''
\newblock in {\em Proceedings of the {{IEEE}}/{{CVF Conference}} on {{Computer Vision}} and {{Pattern Recognition}}}, 2022, pp. 12444--12453.

\bibitem{rathiDIETSNNDirectInput2020a}
Nitin Rathi and Kaushik Roy,
\newblock ``{{DIET-SNN}}: {{Direct Input Encoding With Leakage}} and {{Threshold Optimization}} in {{Deep Spiking Neural Networks}},'' Dec. 2020.

\bibitem{hanRmpsnnResidualMembrane2020a}
Bing Han, Gopalakrishnan Srinivasan, and Kaushik Roy,
\newblock ``Rmp-snn: {{Residual}} membrane potential neuron for enabling deeper high-accuracy and low-latency spiking neural network,''
\newblock in {\em Proceedings of the {{IEEE}}/{{CVF}} Conference on Computer Vision and Pattern Recognition}, 2020, pp. 13558--13567.

\bibitem{zhengGoingDeeperDirectlytrained2021}
Hanle Zheng, Yujie Wu, Lei Deng, Yifan Hu, and Guoqi Li,
\newblock ``Going deeper with directly-trained larger spiking neural networks,''
\newblock in {\em Proceedings of the {{AAAI Conference}} on {{Artificial Intelligence}}}, 2021, vol.~35, pp. 11062--11070.

\bibitem{wuLiafnetLeakyIntegrate2021}
Zhenzhi Wu, Hehui Zhang, Yihan Lin, Guoqi Li, Meng Wang, and Ye~Tang,
\newblock ``Liaf-net: {{Leaky}} integrate and analog fire network for lightweight and efficient spatiotemporal information processing,''
\newblock {\em IEEE Transactions on Neural Networks and Learning Systems}, vol. 33, no. 11, pp. 6249--6262, 2021.

\bibitem{fangIncorporatingLearnableMembrane2021}
Wei Fang, Zhaofei Yu, Yanqi Chen, Timoth{\'e}e Masquelier, Tiejun Huang, and Yonghong Tian,
\newblock ``Incorporating learnable membrane time constant to enhance learning of spiking neural networks,''
\newblock in {\em Proceedings of the {{IEEE}}/{{CVF International Conference}} on {{Computer Vision}}}, 2021, pp. 2661--2671.

\bibitem{liDifferentiableSpikeRethinking2021}
Yuhang Li, Yufei Guo, Shanghang Zhang, Shikuang Deng, Yongqing Hai, and Shi Gu,
\newblock ``Differentiable spike: {{Rethinking}} gradient-descent for training spiking neural networks,''
\newblock {\em Advances in Neural Information Processing Systems}, vol. 34, pp. 23426--23439, 2021.

\bibitem{zhengGoingDeeperDirectlyTrained2020}
Hanle Zheng, Yujie Wu, Lei Deng, Yifan Hu, and Guoqi Li,
\newblock ``Going {{Deeper With Directly-Trained Larger Spiking Neural Networks}},'' Dec. 2020.

\bibitem{buOptimalANNSNNConversion2021}
Tong Bu, Wei Fang, Jianhao Ding, PengLin Dai, Zhaofei Yu, and Tiejun Huang,
\newblock ``Optimal {{ANN-SNN Conversion}} for {{High-accuracy}} and {{Ultra-low-latency Spiking Neural Networks}},''
\newblock in {\em International {{Conference}} on {{Learning Representations}}}, 2021.

\bibitem{zhaiScalingVisionTransformers2022}
Xiaohua Zhai, Alexander Kolesnikov, Neil Houlsby, and Lucas Beyer,
\newblock ``Scaling vision transformers,''
\newblock in {\em Proceedings of the {{IEEE}}/{{CVF Conference}} on {{Computer Vision}} and {{Pattern Recognition}}}, 2022, pp. 12104--12113.

\end{thebibliography}
}

\end{document}